\documentclass{article}
\usepackage{ijcai24}

\usepackage{times}
\usepackage{soul}
\usepackage{url}
\usepackage[hidelinks]{hyperref}
\usepackage[utf8]{inputenc}
\usepackage[small]{caption}
\usepackage{graphicx}
\usepackage{amsmath}
\usepackage{amsthm}
\usepackage{booktabs}
\usepackage{algorithm}
\usepackage{algorithmic}
\usepackage[switch]{lineno}

\usepackage{xcolor}         % colors
\usepackage{multirow}
\usepackage{subfigure}
\usepackage{amssymb}
\usepackage{mathrsfs}
\usepackage{tabularray}

\title{Label Distribution Learning from Logical Label}

\author{
Yuheng Jia$^{1,2}$
\and
Jiawei Tang$^{1,2}$\and
Jiahao Jiang$^{1,2}$
\affiliations
$^1$School of Computer Science and Engineering, Southeast University\\
$^2$Key Laboratory of New Generation Artificial Intelligence Technology and Its Interdisciplinary Applications (Southeast University), Ministry of Education, China
\\
\emails
\{yhjia, jwtang, jhjiang\}@seu.edu.cn
}

\begin{document}

\maketitle

\begin{abstract}
Label distribution learning (LDL) is an effective method to predict the label description degree (a.k.a. label distribution) of a sample. However, annotating label distribution (LD) for training samples is extremely costly. So recent studies often first use label enhancement (LE) to generate the estimated label distribution from the logical label and then apply external LDL algorithms on the recovered label distribution to predict the label distribution for unseen samples. But this step-wise manner overlooks the possible connections between LE and LDL. Moreover, the existing LE approaches may assign some description degrees to invalid labels. To solve the above problems, we propose a novel method to learn an LDL model directly from the logical label, which unifies LE and LDL into a joint model, and avoids the drawbacks of the previous LE methods. We also give the generalization error bound of our method and theoretically prove that directly learning an LDL model from the logical labels is feasible.  Extensive experiments on various datasets prove that the proposed approach can construct a reliable LDL model directly from the logical label, and produce more accurate label distribution than the state-of-the-art LE methods. The code and the supplementary file can be found in https://github.com/seutjw/DLDL.
\end{abstract}

\section{Introduction}
\begin{figure*}[tb]
	\centering
	\subfigure[Natural scene]{
        \label{fig1a}
	\includegraphics[width=0.23\textwidth]{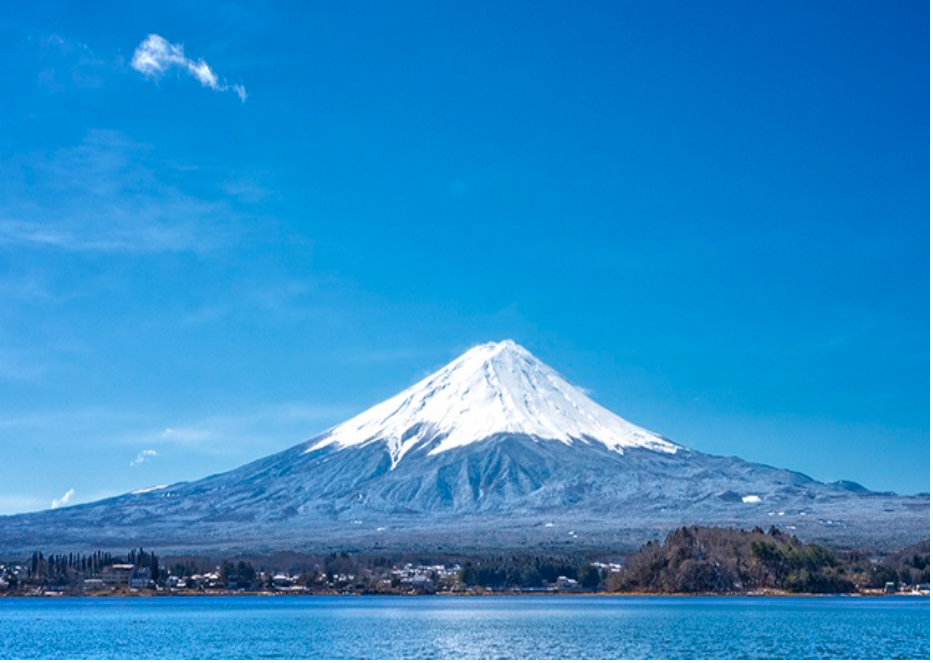}}
	\subfigure[Logical label]{
        \label{fig1b}
	\includegraphics[width=0.23\textwidth]{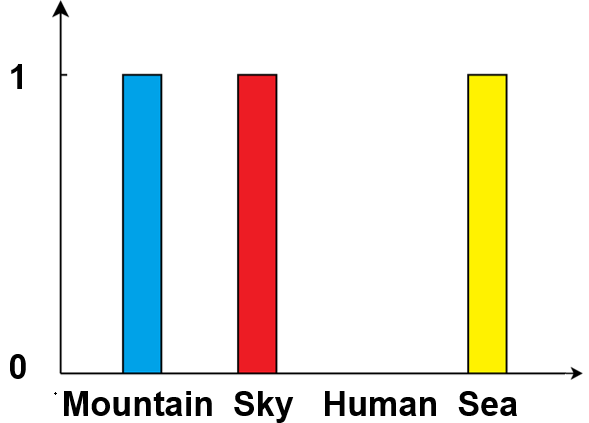}}
        \subfigure[Label distribution]{
        \label{fig1c}
	\includegraphics[width=0.23\textwidth]{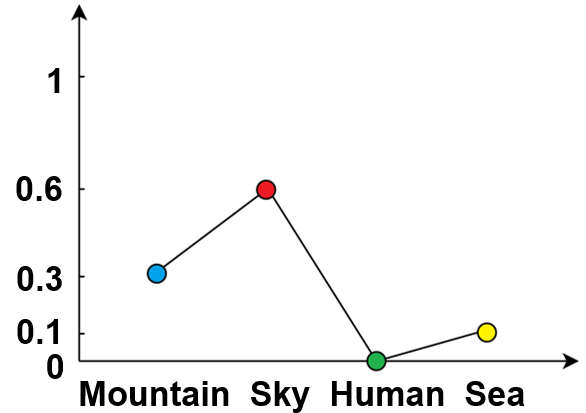}}
	\caption{An example of using label distribution to describe a natural scene image in (a). The histogram in (b) denotes the logical label, indicating whether a label can describe the image in (a), and the line chart in (c) shows the label distribution, revealing to what degree a label can describe the image in (a).}
\end{figure*}

 Multi-label learning (MLL) \cite{MLL1,MLL2} is a well-studied machine learning paradigm where each sample is associated with a set of labels. In MLL, each label is denoted as a logical value (0 or 1), indicating whether a label can describe a sample. However, the logical label cannot precisely describe the relative importance of each label to a sample. To this end, label distribution learning (LDL) \cite{LDL1} was proposed, in which real numbers are used to demonstrate the relative importance of a label to a certain sample. For example, Fig. \ref{fig1a} is a natural scene image, which is annotated with three positive labels (``Sky", ``Mountain" and ``Sea") and one negative label (``Human") as shown in Fig. \ref{fig1b}. Since the relative importance of each label to this image is different, for example, the label ``Mountain" is more important than the label ``Sea", the real numbers (also known as the label description degree) in Fig. \ref{fig1c} can better describe this image. The description degrees of all labels constitute a label distribution (LD). Specifically, if $d^l_{\boldsymbol{x}}$ represents the description degree of the label $l$ to the instance $\boldsymbol{x}$, it's subject to the non-negative constraint $d^l_{\boldsymbol{x}} \in \left[0,1\right]$ and the sum-to-one constraint $\sum_l d^l_{\boldsymbol{x}}=1$. LDL aims to predict the LD for unseen samples, which is a more general paradigm than the traditional MLL.

In LDL, a training set with samples annotated by LDs is required to train an LDL model. Unfortunately, the acquisition of label distributions of instances is a very costly and time-consuming process. Moreover, in reality, most datasets are only annotated by logical labels, which cannot be used directly by the existing LDL methods. Then a question naturally arises: \textit{can we directly train an LDL model from the logical labels?} 

Recently, label enhancement (LE) \cite{LE1} was proposed to partially answer this question. Specifically, LE first recovers the label distributions of training samples from the logical labels and then performs an existing LDL algorithm on the recovered LDs. Following \cite{LE1}, many variants of LE were proposed, such as \cite{LEMLL}, \cite{LESC} and \cite{LP}. We refer the readers to the related work section for more details. To achieve LE, those methods generally construct a linear mapping from the features to the logical labels directly, and then normalize the output of the linear mapping as the recovered LDs. However, those LE methods may assign positive description degrees to some negative logical labels. Moreover, as all the positive logical labels are annotated as $1$, simply building a mapping from features to the logical labels is not the best choice, which will not differentiate the description degrees of different labels. Last but not least, the LE-based strategy is a step-wise manner to predict LD for unseen samples, which may lose the connection between LE and LDL model training. 

In this paper, we come up with a novel model named DLDL, i.e., Directly Label Distribution Learning from the logical label, which gives a favorable answer to the question ``\textit{can we directly train an LDL model from the logical labels}". The major contributions of our method are summarized as follows:

\begin{itemize}
\item  Our model combines LE and LDL into a single model, which can learn an LDL model directly from the instances annotated by the logical labels. By the joint manner, the LE and LDL processes will better match each other. 

\item We strictly constrain the description degree $d^l_{\boldsymbol{x}}$ to be 0 when the logical value corresponding to the label $l$ is 0. The constraint will avoid assigning positive description degrees to the negative logical labels.

\item The existing LE methods usually minimize a least squares loss between the recovered LD and the logical label, which can not well differentiate the description degrees of different labels. The proposed model discards this fidelity term, and uses KL-divergence to minimize the difference between the recovered LD and the predictive LD, which is a better difference measure for two distributions.

\item By using the Rademacher complexity, we show the generalization bound of the proposed model and theoretically prove that it is possible to construct an LDL model directly from the logical labels for the first time.
\end{itemize}

Extensive experiments on six benchmark datasets clearly show that the LD recovered by our method is better than that recovered by state-of-the-art LE methods on the training set, and the prediction performance of our model on the testing set is also better than the traditional step-wise strategy.

\section{Related Works}
\textbf{Notations}: Let $n$, $m$ and $c$ represent the number of samples, the dimension of features, and the number of labels. Let $\boldsymbol{x}\in\mathbb{R}^m$ denote a feature vector and $\boldsymbol{y}\in\left\{0,1\right\}^c$ denote its corresponding logical label vector. The feature matrix and the corresponding logical label matrix can be denoted by $\mathbf{X}=\left[\boldsymbol{x}_1;\boldsymbol{x}_2;\ldots;\boldsymbol{x}_n\right]\in\mathbb{R}^{n\times m}$ and $\mathbf{Y}=\left[\boldsymbol{y}_1;\boldsymbol{y}_2;\ldots;\boldsymbol{y}_n\right]\in\left\{0,1\right\}^{n\times c}$, respectively. Let $\mathcal{Y}=\left\{l_{1}, l_{2}, \ldots, l_{c}\right\}$ be the complete set of labels. The description degree of the label $l$ to the instance $\boldsymbol{x}$ is denoted by $d^l_{\boldsymbol{x}}$, which satisfies $d^l_{\boldsymbol{x}}\in\left[0,1\right]$ and $\sum_l d^l_{\boldsymbol{x}}=1$, and the label distribution of $\boldsymbol{x}_i$ is denoted by $\boldsymbol{d}_i=\left(d^{l_{1}}_{\boldsymbol{x}_i},d^{l_{2}}_{\boldsymbol{x}_i},\ldots,d^{l_{c}}_{\boldsymbol{x}_i}\right)$.

\subsection{Label Distribution Learning}
LDL is a new machine learning paradigm that constructs a model to predict the label distribution of samples. At first, LDL were achieved through problem transformation that transforms LDL into a set of single label learning problems such as PT-SVM, PT-Bayes \cite{LDL1}, or through  algorithm adaptation that adopts the existing machine learning algorithms to LDL, such as AA-kNN and AA-BP \cite{LDL1}. SA-IIS \cite{LDL1} is the first model that specially designed for LDL, whose objective function is a mixture of maximum entropy loss \cite{maxentropymodel} and KL-divergence. Based on SA-IIS, SA-BFGS \cite{LDL1} adopts BFGS to optimize the loss function, which is faster than SA-IIS. Sparsity conditional energy label distribution learning (SCE-LDL) \cite{SCELDL} is a three-layer energy-based model for LDL. In addition, SCE-LDL is improved by incorporating sparsity constraints into the objective function. To reduce feature noise, latent semantics encoding for LDL (LSE-LDL) \cite{LSELDL} converts the original data features into latent semantic features, and removes some irrelevant features by feature selection. LDL forests (LDLFs) \cite{LDLFs} is based on differentiable decision trees and may be combined with representation learning to provide an end-to-end learning framework. LDL by optimum transport (LDLOT) \cite{LDLOT} builds an objective function using the optimal transport distance measurement and label correlations. $\textit{L}^2$ \cite{L2} is the first attempt to combine LE and LDL into a unified model, utilizing the manifold structure of samples and label correlations to learn the prediction model.

\subsection{Label Enhancement}
The above mentioned LDL methods all assume that in the training set, each sample is annotated with label distribution. However, precisely annotating the LDs for the training samples is extremely costly and time-consuming. On the contrary, many datasets annotated by logical labels are readily available. To this end, LE was proposed, which aims to convert the logical labels of samples in training set to LDs. GLLE \cite{LE1} is the first LE algorithm, which assumes that the LDs of two instances are similar to each other if they are similar in the feature space. LEMLL \cite{LEMLL} adopts the local linear embedding technique to evaluate the relationship of samples in the feature spaces. Generalized label enhancement with sample correlations (gLESC) \cite{LESC} tries to obtain the sample correlations from both of the feature and the label spaces. Bidirectional label enhancement (BD-LE) \cite{BD-LE} takes the reconstruction errors from the label distribution space to the feature space into consideration.

All of these LE methods and LE loss of $\textit{L}^2$ \cite{L2} can be generally formulated as
\begin{equation}
\small
\label{generalLE}
    \min_{\mathbf{W}}\vert\vert\mathbf{XW}-\mathbf{Y}\vert\vert_F^2+\phi (\mathbf{XW},\mathbf{X}), 
\end{equation}
where $\mathbf{X}$ and $\mathbf{Y}$ are the feature matrix and logical label matrix, $\|\cdot\|_F$ denotes the Frobenius of a matrix, $\mathbf{W}\in\mathbb{R}^{m\times c}$ builds a linear mapping from features to logical labels, and $\phi (\mathbf{XW},\mathbf{X})$ models the geometric structure of samples, which is used to assist LD recovery. After minimizing Eq. (\ref{generalLE}), those methods usually normalize $\mathbf{XW}$ as the recovered LD.

Although those LE methods have achieved great success, they still suffer from the following \textit{limitations}. Firstly, the fidelity term that minimizes the distance between the recovered LD (i.e., $\mathbf{XW}$) and the logical label $\mathbf{Y}$ is inappropriate because the logical labels are annotated as the same value, and the linear mapping will not differentiate the description degrees for different labels. Besides, the Frobenius norm is also not the best choice to measure the difference between two distributions. Secondly, Eq. (\ref{generalLE}) doesn't consider the physical restriction of the label distribution, i.e., $\forall i, \textbf{0} \leq \boldsymbol{d}_i \leq \boldsymbol{y}_i, \sum_l d^l_{\boldsymbol{x}_i}=1$. Although those methods perform a post-normalization to satisfy those constraints, they may assign positive description degrees to some negative logical labels. Furthermore, to predict the LD for unseen samples, those methods need to first perform LE and then carry out an existing LDL algorithm. The step-wise manner does not consider potential connections between LE and LDL.

\section{The Proposed Approach}
To solve the above-mentioned issues, we propose a novel model named DLDL. Different from the previous two-step strategy, our method can generate a label distribution matrix $\mathbf{D} = \left[\boldsymbol{d}_1;\boldsymbol{d}_2;\ldots;\boldsymbol{d}_n\right] \in\mathbb{R}^{n\times c}$ for samples annotated by logical labels and at the same time construct a weight matrix $\mathbf{W}\in\mathbb{R}^{m\times c}$ to predict LD for unseen samples.

We use the following priors to recover the LD matrix $\mathbf{D}$ from the logical $\mathbf{Y}$. First, as each row of $\mathbf{D}$ (e.g., $\boldsymbol{d}_i$) denotes the recovered LD for a sample, it should obey the non-negative constraint and the sum-to-one constraint to make it a well-defined distribution, i.e., $\mathbf{0}_c\leq\boldsymbol{d}_i\leq \mathbf{1}_c$ and $\boldsymbol{d}_i\mathbf{1}_c=1$, where $\mathbf{0}_c$, $\mathbf{1}_c\in \mathbb{R}^c$ denote an all-zeros vector and an all-ones vector, and $\mathbf{0}\leq\boldsymbol{d}_i\leq \mathbf{1}$ means each element of $\boldsymbol{d}_i$ is non-negative and no more than $1$. Moreover, to avoid assigning a positive description to a negative logical label, we require that $\mathbf{0}\leq \boldsymbol{d}_i\leq \boldsymbol{y}_i, \forall i$. Reformulating the above constraints in the matrix form, we have $\mathbf{0}_{m\times c}\leq \mathbf{D}\leq \mathbf{Y},$ $\mathbf{D}\mathbf{1}_c = \mathbf{1}_n$, where $\mathbf{0}_{m\times c}$ is an all-zeros matrix with size $m\times c$, and the above inequalities are held element-wisely.

Second, we utilize the geometric structure of the samples to help recover the LDs from the logical labels, i.e., if two samples are similar to each other in the feature space, they are likely to own the similar LDs. To capture the similarity among samples, we define the local similarity matrix $\mathbf{A}=\left[A_{i j}\right]_{n \times n}$ as
\begin{equation}
\small
    A_{i j}=
    \begin{aligned}
    \begin{cases}
    \exp \left(-\left\|\boldsymbol{x}_i-\boldsymbol{x}_j\right\|_{2}^{2} / 2\sigma^{2}\right),&\text{if }\boldsymbol{x}_j \in \mathcal{N}\left(\boldsymbol{x}_{i}\right)\\
    0,\quad &\text{otherwise}.
    \end{cases}
    \end{aligned}
\end{equation}
$\mathcal{N}\left(\boldsymbol{x}_{i}\right)$ denotes the $k$-nearest neighbors of $\boldsymbol{x}_{i}$, and $\sigma$ is the hyper-parameter of the RBF kernel. Based on the constructed local similarity matrix $\mathbf{A}$, we can infer that the value of $\vert\vert \boldsymbol{d}_{i}-\boldsymbol{d}_{j} \vert\vert_2$ is small when $A_{ij}$ is large, i.e.,
\begin{equation}
\small
\label{lambda}
    \begin{aligned}
         \min_{\mathbf{D}} \ \underset{i,j}{\sum}A_{ij}\vert\vert \boldsymbol{d}_{i}-\boldsymbol{d}_{j}\vert\vert_{2}^{2} =\rm{tr}\left(\mathbf{D}^T\mathbf{G}\mathbf{D}\right),\\
    \end{aligned}
\end{equation}
in which $\rm{tr}\left( \cdot \right)$ denotes the trace of a matrix, and $\mathbf{G}=\hat{\mathbf{A}}-\mathbf{A}$ is the graph Laplacian matrix \cite{Graph1,Graph2} and $\hat{\mathbf{A}}$ is a diagonal matrix with the elements $\hat{A}_{i i}=\sum_{j=1}^{n} A_{i j}$.

By taking the above priors into consideration, the LD matrix $\mathbf{D}$ can be inferred from the logical label matrix $\mathbf{Y}$, i.e.,
\begin{equation}
\small
\begin{aligned}
&\underset{\mathbf{D}}{\min} \ \rm{tr}\left(\mathbf{D}\mathbf{G}\mathbf{D}^T\right)+\vert\vert\mathbf{D}\vert\vert_{F}^2 \qquad \\
&\text{s.t.} \ \textbf{0}_{n\times c} \leq \mathbf{D} \leq \mathbf{Y}, \mathbf{D}\mathbf{1}_c=\mathbf{1}_n,
\end{aligned}
\label{LEObject}
\end{equation}
where an additional term $\|\mathbf{D}\|_{\rm{F}}^2$ is imposed as regularization. Different from the previous LE methods in Eq. (\ref{generalLE}), we discard the fidelity term $\|\mathbf{Y}-\mathbf{WX}\|_{\rm{F}}^2$. But as we strictly constrain that $\mathbf{0}\leq\mathbf{D}\leq\mathbf{Y}$, it will have a similar effect as the above fidelity term, i.e., encouraging the description degree of the valid labels to be large. But our approach avoids the drawbacks of the fidelity term, i.e., assigning label distribution degree to some negative logical labels and  not differentiating the description degrees of different labels.

Based on the recovered LD, we adopt a non-linear model that maps the features to the recovered LD, which can be used to predict the LD for unseen samples. Specifically, the non-linear model is formulated as
\begin{equation}
\small
\label{P}
    P_{ij}=\dfrac{1}{Z} \exp \left(\sum^m_{k=1} X_{ik}W_{kj}\right),
\end{equation}
where $\mathbf{W}$ is the weight matrix, $\mathbf{P}=[P_{ij}]_{n\times c}$ is the prediction matrix, and $Z=\sum^c_{j=1} \exp \left(\sum^m_{k=1} X_{ik}W_{kj} \right)$ is a normalization term.  To infer the weight matrix $\mathbf{W}$, we minimize the KL-divergence between the recovered label distribution matrix $\mathbf{D}$ and the prediction matrix $\mathbf{P}$, because KL-divergence is a good metric to measure the difference between the two distributions. Accordingly, the loss function for inferring $\mathbf{W}$ becomes
\begin{equation}
\small
\begin{aligned}
\underset{\mathbf{W}}{\min} \
\rm{KL}(\mathbf{D},\mathbf{P})+\vert\vert\mathbf{W}\vert\vert_{F}^2,
\end{aligned}
\label{LDLObject}
\end{equation}
where the widely-used squared Frobenius norm is imposed to control the model complexity.

To directly induce an LD prediction model from the logical labels, we combine Eqs. (\ref{LEObject}) and (\ref{LDLObject}) together and the final optimization problem of our method becomes:
\begin{equation}
\small
\begin{aligned}
&\underset{\mathbf{D},\mathbf{W}}{\min} \
\rm{KL}(\mathbf{D},\mathbf{P})+\alpha tr\left(\mathbf{D}\mathbf{G}\mathbf{D}^T\right)+\beta\vert\vert\mathbf{D}\vert\vert_{F}^2+\gamma\vert\vert\mathbf{W}\vert\vert_{F}^2 \\
& \text{s.t.} \ \textbf{0}_{n\times c} \leq \mathbf{D} \leq \mathbf{Y}, \mathbf{D}\mathbf{1}_c=\mathbf{1}_n,
\end{aligned}
\label{finalObject}
\end{equation}
in which $\alpha$, $\beta$ and $\gamma$ are hyper-parameters to adjust the contributions of different terms. By minimizing Eq. (\ref{finalObject}), the proposed model can recover the LD for samples with logical labels, and simultaneously, can predict LD for unseen samples through Eq. (\ref{P}). 

\subsection{Optimization}
Eq. (\ref{finalObject}) has two variables, we solve it by an alternative and iterative process, i.e., update $\mathbf{W}$ with fixed $\mathbf{D}$, and then update $\mathbf{D}$ with fixed $\mathbf{W}$. 
%\subsubsection{Update W} 

\textbf{1) Update W} When $\mathbf{D}$ is fixed, the optimization problem (\ref{finalObject}) with respect to $\mathbf{W}$ is formulated as
\begin{equation}
\small
\label{updateW}
    \begin{aligned}
        \underset{\mathbf{W}}{\min} \ &\rm{KL}(\mathbf{D},\mathbf{P})+\gamma\vert\vert\mathbf{W}\vert\vert_{F}^2.
    \end{aligned}
\end{equation}
Expanding the KL-divergence and substituting $P_{ij}$ in Eq. (\ref{P}) into Eq. (\ref{updateW}), we obtain the objective function of $\mathbf{W}$ as
\begin{equation}
\small
\begin{aligned}
\label{TW}
    T(\mathbf{W})=&
    \sum_{i} {\rm ln} \sum_{j}\exp \left(\sum_{k} X_{i k} W_{k j}\right) + \gamma\vert\vert\mathbf{W}\vert\vert_{\rm{F}}^2 \\
    &- \sum_{i,j} D_{i j}\sum_{k} X_{i k} W_{k j}.
\end{aligned}
\end{equation}
Eq. (\ref{TW}) is a convex problem, we use gradient descent algorithm to update $\mathbf{W}$, where the gradient is expressed as
\begin{equation}
\small
\label{gradient_W}
\begin{aligned}
    \dfrac{\partial \ T\left( \mathbf{W} \right)}{\partial \ W_{k j}} =& \sum_{i}\dfrac{\exp\left(\sum_{k} X_{i k} W_{k j}\right) X_{i k}}{\sum_{j}\exp\left(\sum_{k}X_{i k} W_{k j}\right)} - \sum_{i} D_{i j} X_{i k}\\
    &+2\gamma W_{k j}.
\end{aligned}
\end{equation}

\begin{table*}[t]
\centering
	\begin{minipage}[t]{0.47\linewidth}
        \begin{algorithm}[H]
        \small
        \caption{Solve the Problem in (\ref{Dorigin})}
        \label{alg1}
        \textbf{Input}: The initial label distribution matrix $\mathbf{D}$, the initial weight matrix $\mathbf{W}$, and $\mathbf{B}=\mathbf{\Phi}=\mathbf{0}_{n\times c}$.\\
        \textbf{Parameter}: $\rho=1.2$, $\tau=0.001$ and $\tau_{max}=0.002$.\\
        \mbox{\textbf{Output}: The updated $\mathbf{D}$.}
        \begin{algorithmic}[1] %[1] enables line numbers
        \WHILE{$\vert\vert\mathbf{B}-\mathbf{D}\vert\vert_{\infty}>10^{-3}$}
        \STATE Update $\mathbf{D}$ by Eq. (\ref{updateD});
        \STATE Update $\mathbf{B}$ by solving Eq. (\ref{updateB});
        \STATE Update $\mathbf{\Phi}=\mathbf{\Phi}+\tau(\mathbf{B}-\mathbf{D})$ and $\tau=\min(\rho\tau,\tau_{max})$;
        \ENDWHILE 
        \STATE $\mathbf{D}=\mathbf{B}$; \textbf{return} $\mathbf{D}$.
        \end{algorithmic}
        \end{algorithm}
	\end{minipage}
    \hfill
	\begin{minipage}[t]{0.47\linewidth}
        \begin{algorithm}[H]
        \small
        \caption{The DLDL Algorithm}
        \label{alg2}
        \textbf{Input}: The feature matrix $\mathbf{X}$ and the logical label matrix $\mathbf{Y}$.\\
        \textbf{Parameter}: $\alpha,\beta,\lambda$ and $\sigma$.\\
        \textbf{Output}: The label distribution matrix $\mathbf{D}$ and the weight matrix $\mathbf{W}$.
        \begin{algorithmic}[1] %[1] enables line numbers
        \STATE Initialize $\mathbf{W}$ and $\mathbf{D}$;
        \REPEAT
        \STATE Update $\mathbf{W}$ by using Eq. (\ref{gradient_W});
        \STATE Update $\mathbf{D}$ by Algorithm \ref{alg1};
        \UNTIL{convergence or reaches maximum number of iterations;}
        \STATE \textbf{return} $\mathbf{D}$ and $\mathbf{W}$.
        \end{algorithmic}
        \end{algorithm}
	\end{minipage}

\end{table*}

%\subsubsection{Update D}

\textbf{2) Update D} With fixed $\mathbf{W}$, the optimization problem (\ref{finalObject}) regarding $\mathbf{D}$ is formulated as
\begin{equation}
\small
\label{Dorigin}
\begin{aligned}
& \underset{\mathbf{D}}{\min} \ \rm{KL}(\mathbf{D},\mathbf{P})+\alpha \rm{tr}\left(\mathbf{D}\mathbf{G}\mathbf{D}^T\right)+\beta\vert\vert\mathbf{D}\vert\vert_{F}^2 \\
& \text {s.t.} \ \textbf{0} \leq \mathbf{D} \leq \mathbf{Y}, \mathbf{D}\mathbf{1}_c=\mathbf{1}_n.
\end{aligned}
\end{equation}
Eq. (\ref{Dorigin}) involves multiple constraints and diverse losses, we introduce an auxiliary variable $\mathbf{B} = \mathbf{D} \in \mathbb{R}^{n\times c}$ to simplify it, and the corresponding augmented Lagrange equation becomes
\begin{equation}
\small
\label{lagrangeD}
\begin{aligned}
& \underset{\mathbf{D},\mathbf{B}}{\min} \ \rm{KL}(\mathbf{D},\mathbf{P})+\alpha\rm{tr}\left(\mathbf{D}^T\mathbf{G}\mathbf{D}\right) +\beta\vert\vert\mathbf{D}\vert\vert_{F}^2 \\
& + \left<\mathbf{\Phi},\mathbf{B}-\mathbf{D}\right>+\frac{\tau}{2}\vert\vert\mathbf{B}-\mathbf{D}\vert\vert^{2}_{F} \\
& \text {s.t.} \ \textbf{0} \leq \mathbf{B} \leq \mathbf{Y}, \mathbf{B}\mathbf{1}_c=\mathbf{1}_n,
\end{aligned}
\end{equation}
where $\mathbf{\Phi}\in \mathbb{R}^{n\times c}$ is the Lagrange multiplier, and $\tau$ is parameter to introduce the augmented equality constraint. Eq. (\ref{lagrangeD}) can be minimized by solving the following sub-problems iteratively.

\textbf{\textit{a) $\mathbf{D}$-subproblem:}} Removing the irrelated terms regarding $\mathbf{D}$, the $\mathbf{D}$-subproblem of Eq. (\ref{lagrangeD}) becomes
\begin{equation}
\small
\begin{aligned}
\label{Dsubproblem}
\underset{\mathbf{D}}{\min} \ U\left( \mathbf{D} \right) &= \rm{KL}(\mathbf{D},\mathbf{P}) +\alpha \rm{tr}\left(\mathbf{D}^T\mathbf{G}\mathbf{D}\right) +\beta\vert\vert\mathbf{D}\vert\vert_{F}^2 \\
&+ \left<\mathbf{\Phi},\mathbf{B}-\mathbf{D}\right>+\frac{\tau}{2}\vert\vert\mathbf{B}-\mathbf{D}\vert\vert^{2}_{F},
\end{aligned}
\end{equation}
then its gradient regarding $\mathbf{D}$ can be expressed as:
\begin{equation}
\small
\label{updateD}
\begin{aligned}
\dfrac{\partial \ U\left( \mathbf{D} \right)}{\partial \ D_{i j}}=
    \begin{cases}
    1+\ln{D_{i j}}-\ln{P_{i j}}+\alpha(\mathbf{G}^T+\mathbf{G})\mathbf{D}
    \\+2\beta\mathbf{D}-\Phi+\tau(\mathbf{D}-\mathbf{B}),&\text{if }D_{i j}\neq0\\
    0,&\text{if }D_{i j}=0.\\
    \end{cases}
\end{aligned}
\end{equation}
\\
\textbf{\textit{b) $\mathbf{B}$-subproblem:}} When $\mathbf{D}$ is fixed, the problem (\ref{lagrangeD}) can be written as
\begin{equation}
\small
\label{Borigin}
\begin{aligned}
&\underset{\mathbf{B}}{\min} \ \left<\mathbf{\Phi},\mathbf{B}-\mathbf{D}\right>
+\frac{\tau}{2}\vert\vert\mathbf{B}-\mathbf{D}\vert\vert^{2}_{F} \\
&\text {s.t.} \ \textbf{0} \leq \mathbf{B} \leq \mathbf{Y}, \mathbf{B}\mathbf{1}_c=\mathbf{1}_n.
\end{aligned}
\end{equation}
Let $\hat{\boldsymbol{b}}=[\boldsymbol{b}_1^T;\boldsymbol{b}_2^T;\cdots;\boldsymbol{b}_n^T]\in\mathbb{R}^{nc}=vec(\mathbf{B})$, where $\boldsymbol{b}_i$ is the $i$-th row of $\mathbf{B}$ and $vec(\cdot)$ is the vectorization operator. Then, the problem (\ref{Borigin}) is equivalent to:
\begin{equation}
\small
\label{updateB}
    \begin{aligned}
        &\underset{\hat{\boldsymbol{b}}} {\min} \ \hat{\boldsymbol{b}}^T \mathbf{\Sigma} \hat{\boldsymbol{b}} - \left(2\hat{\boldsymbol{d}}^T+\frac{2}{\tau} \hat{\boldsymbol{\phi}}^T\right) \hat{\boldsymbol{b}} \\
        &\text {s.t.} \ \mathbf{0}_{nc} \leq \hat{\boldsymbol{b}} \leq \hat{\boldsymbol{y}}, \sum_{j=c(i-1)+1}^{ci}\hat{\boldsymbol{b}}_j=1 \ \left (\forall \ 0\leq i\leq n \right),
    \end{aligned}
\end{equation}
in which $\hat{\boldsymbol{d}}$, $\hat{\boldsymbol{\phi}}$ and $\hat{\boldsymbol{y}}$ represent $vec(\mathbf{D})$, $vec(\mathbf{\Phi})$ and $vec(\mathbf{Y})$, respectively, and $\hat{\boldsymbol{b}}_j$ is the $j$-th element of $\hat{\boldsymbol{b}}$. Eq. (\ref{updateB}) is a quadratic programming (QP) problem that can be solved by off-the-shelf QP tools. 

The detailed solution to update $\mathbf{D}$ in Eq. (\ref{Dorigin}) is summarized in Algorithm \ref{alg1} and the pseudo code of DLDL is finally presented in Algorithm \ref{alg2}. In this paper, we initialize $\mathbf{W}$ as an identity matrix and initialize $\mathbf{D}$ according to the method in \cite{FLE}.

%\subsubsection{Initialization of W and D}

%The above alternative updating of $\mathbf{W}$ and $\mathbf{D}$ needs an initialization of them. As is suggested by \cite{FLE}, we initialize $\mathbf{W}$ as an identity matrix and we initialize $\mathbf{D}$ by the pseudo label distribution matrix in \cite{FLE}.

\subsection{Generalization Bound}
In this section, we first provide Rademacher complexity for DLDL, which is a commonly used tool for comprehensive analysis of data-dependent risk bounds.

\textbf{Definition 1. } \textit{Let } $\mathcal{H}$ \textit{ be a family of functions mapping from } $\mathcal{X}$\textit{ to [0,1] and } $\mathcal{S}$ \textit{be a set of fixed samples with size }$n$\textit{. Then, the empirical Rademacher complexity of } $\mathcal{H}$\textit{ with respective to }$\mathcal{S}$\textit{ is defined as}
\begin{equation}
\small
\widehat{\mathcal{R}}_{S}(\mathcal{H})=\mathbb{E}_{\boldsymbol{\sigma}}\left[\sup _{h \in \mathcal{H}} \frac{1}{n} \sum_{i=1}^{n} \sigma_{i} h\left(x_{i}\right)\right].
\end{equation}

\textbf{Lemma 1. } \textit{Let $\mathcal{H}$ be a family of functions. For a loss function $\ell$ upper bounded by $\Theta$, then for any $\delta>0$, with probability at least $1-\delta$, for all $h \in \mathcal{H}$ such that
\begin{equation}
\small
\label{Lemma1}
    \begin{aligned}
        \mathcal{L}(h) \leq \mathcal{L}_{S}(h)+\widehat{\mathcal{R}}_{S}(\ell \circ \mathcal{H})+3 \Theta \sqrt{\frac{\log 2 / \delta}{2 n}},
    \end{aligned}
\end{equation}
where $\mathcal{L}(h)$ and $\mathcal{L}_{S}(h)$ are the generalization risk and empirical risk with respective to h.}

\textbf{Theorem 1. } \textit{The KL-divergence loss function $\ell$ can be written as $\rm{KL}(\mathbf{D},\mathbf{P})$, where $\mathbf{D}\in\mathbb{R}^{n\times c}$ and $\mathbf{P}\in\mathbb{R}^{n\times c}$ are the recovered LD matrix and the prediction matrix respectively, in which $\mathbf{P}$ can be expressed by the prediction weight matrix $\mathbf{W}\in\mathbb{R}^{m\times c}$. Let $\mathcal{H}=\mathbf{D}\times\mathbf{W}$ represent the family of functions for DLDL, with functions $(\mathbf{D},\mathbf{W})\in\mathcal{H}$. Our algorithm further encourages the complexity of $\mathbf{W}$ and the rank of $\mathbf{D}$ are upper bounded by $\epsilon_1$ and $\epsilon_2$ respectively, i.e., $\vert\vert\mathbf{W}\vert\vert_F\leq\epsilon_1$ and $rank(\mathbf{D})\leq\epsilon_2$. According to \textbf{Definition 1}, the Rademacher complexity of DLDL with KL-divergence loss $\ell$ is upper bounded as follows:}

\begin{equation}
\small
\label{RS}
    \begin{aligned}
        \widehat{\mathcal{R}}_{S}(\ell \circ \mathcal{H})\leq \dfrac{\epsilon_2\sqrt{cm\cdot {\exp{(m\vert X_{max}\epsilon_1}\vert)}/{\exp{(-m\vert X_{min}\epsilon_1\vert)}}}}{\sqrt{n}},
    \end{aligned}
\end{equation}
\textit{in which $X_{max}$, $X_{min}$ are the maximum and minimum element in the feature matrix $\mathbf{X}$ respectively.}

The proof is given in section \textbf{A} of the supplementary file. According to \textbf{Lemma 1} and \textbf{Theorem 1}, the Rademacher complexity of DLDL is controlled by the number of training samples  $n$. Because the numerator is a finite number, when DLDL has infinite training samples ($n$ tends to infinity), its Rademacher complexity tends to zero.  

\textbf{Theorem 2. } \textit{Denote $\mathbf{D}\in\mathbb{R}^{n\times c}$ and $\mathbf{W}\in\mathbb{R}^{m\times c}$ as the recovered LD matrix and the prediction weight matrix, and the complexity of $\mathbf{W}$ and the rank of $\mathbf{D}$ are upper bounded by $\epsilon_1$ and $\epsilon_2$ respectively. Then we have the upper bound of $\Theta$:}

\begin{equation}
\small
\label{theta}
    \begin{aligned}
    \Theta\leq\Sigma_{i=1}^n\Sigma_{j=1}^c\ln\frac{{(m\exp{(m\vert X_{max}\epsilon_1}\vert))}}{{\exp{(-m\vert X_{min}\epsilon_1\vert)}}}.
    \end{aligned}
\end{equation}

The proof is given in section \textbf{A} of the supplementary file. The right side of Eq.\eqref{theta} is also a finite number, so when $n$ tends to infinity, the term $3 \Theta \sqrt{\frac{\log 2 / \delta}{2 n}}$ in \textbf{Lemma 1} tends to zero.

According to \textbf{Lemma 1}, \textbf{Theorem 1} and \textbf{Theorem 2}, we finally have 

\begin{equation}
\small
\label{proof_final}
    \begin{aligned}
        \mathcal{L}(h) \leq \mathcal{L}_{S}(h)+\dfrac{\epsilon}{\sqrt{n}},
    \end{aligned}
\end{equation}
in which $\epsilon$ is a finite number. From Eq.\eqref{proof_final}, we can clearly see that when the number of training samples tends to infinity, the generalization risk of $h$ will be upper-bounded by the empirical risk of it.

The above generalized error bound is based on that the recovered label distribution matrix $\mathbf{D}$ is just the ground-truth one, so the last thing to prove is that under some certain conditions, DLDL can recover ground-truth label distribution from logical label.

\textbf{Theorem 3. } \textit{If the number of zeros in the ground-truth label distribution matrix $\mathbf{D}_g$ is large enough, then the average difference between $\mathbf{D}$ and $\mathbf{D}_g$ has an upper bound:}

\begin{equation}
\small
\label{LEproof}
    \begin{aligned}
        \frac{\sum_{i,l}\vert d_{il}-d_{g(il)}\vert}{nc} \leq\frac{2p_2\epsilon}{(1+\epsilon)+\frac{\lambda}{\Sigma k}}+p_3,
    \end{aligned}
\end{equation}
\textit{in which $p_2\in[0,1]$ and $p_3$, $\epsilon$, $\frac{\lambda}{\Sigma k}$ are small positive numbers.}

The detailed proof of \textbf{Theorem 3} is given in section \textbf{B} of the supplementary file. The average difference between $\mathbf{D}$ and $\mathbf{D}_g$ tends to zero when $p_3$, $\epsilon$, $\frac{\lambda}{\Sigma k}$ are small enough, which indicates that DLDL can precisely recover the ground-truth label distribution from logical label under this circumstance. \textit{Combining all the lemmas and theorems together, we conclude that directly learning label distribution learning from logical label is theoretically feasible.}

\section{Experiments}
%\subsection{Real-World Datasets}
We select six real-world datasets from various fields for experiment. Natural Scene (abbr. NS) \cite{LDL1,NS} is generated from the preference distribution of each scene image, SCUT-FBP (abbr. SCUT) \cite{SCUTFBP} is a benchmark dataset for facial beauty perception, RAF-ML (abbr. RAF) \cite{RAFML} is a multi-label facial expression dataset, SCUT-FBP5500 (abbr. FBP) \cite{FBP5500} is a big dataset for facial beauty prediction, Ren\_CECps (abbr. REN) \cite{Ren-Cecps} is a Chinese emotion corpus of weblog articles, and Twitter\_LDL (abbr. Twitter) \cite{twitter} is a visual sentiment dataset. Some basic statistics about these datasets are given in section \textbf{C} of the supplementary file.

To verify whether our method can directly learn an LDL model from the logical labels, we generate the logical labels from the ground-truth LDs. Specifically, when the description degree is higher than a predefined threshold $\delta$, we set the corresponding logical label to 1; otherwise, the corresponding logical label is 0. In this paper, $\delta$ is fixed to 0.01.

\subsection{Baselines and Settings}
In this paper, we split each dataset into three subsets: training set (60\%), validation set (20\%) and testing set (20\%). The training set is used for recovery experiments, that is, we perform LE methods to recover the LDs of training instances; the validation set is used to select the optimal hyper-parameters for each method; the testing set is used for predictive experiments, that is, we use an LDL model learned from the training set with the recovered LDs to predict the LDs of testing instances.

In the recovery experiment, for DLDL, $\alpha$ and $\gamma$ are chosen among \{$10^{-3},10^{-2},\cdots,10,10^2$\}, $\beta$ is selected from \{$10^{-3},10^{-2},\cdots,1,10$\}, the maximum of iterations $t$ is fixed to 5, the number of neighbors $k$ is set to 20. We compare DLDL with five state-of-the-art LE methods and a unified method, each configured with suggested configurations in respective literature: 
\textbf{1)} FLE \cite{FLE}: $\alpha$, $\beta$, $\lambda$, $\gamma$ are chosen among \{$10^{-3},10^{-2},\cdots,10,10^2$\};
\textbf{2)} GLLE \cite{LE1}: $\lambda$ is chosen among \{$10^{-3},10^{-2},\cdots,1$\};
\textbf{3)}: LEMLL \cite{LEMLL}: the number of neighbors $K$ is set to 10 and $\epsilon$ is set to 0.2;
\textbf{4)} LESC \cite{LESC}: $\beta$ is set to 0.001 and $\gamma$ is set to 1;
\textbf{5)} FCM \cite{FCM}: the number of clustering centers is set to 10 times of the number of labels.
\textbf{6)} $\textit{L}^2$ \cite{L2}: $\alpha$, $\gamma$, $\lambda$ are chosen among \{$10^{-5},10^{-3},\cdots,10^3,10^5$\}, $k$ is set to 20 and $\beta$, $\mu$ keep the default value.

In the predictive experiment, we apply SA-BFGS \cite{LDL1} algorithm to predict the LDs of testing instances based on the recovered LDs by performing the five baseline LE methods. Note that DLDL and $\textit{L}^2$ can directly predict the LDs without an external LDL model.

As suggested in \cite{LDL1}, we adopt three distance metrics (i.e., Chebyshev, Clark and One-error) and one similarity metric (i.e., Intersection) to evaluate the recovery and the predictive performances. The formulas of these evaluation metrics are summarized in section \textbf{C} of the supplementary file.

\subsection{Visualization of the Recovered Label Distributions}
\begin{figure}[!ht]
    \centering
    \subfigure{
    \includegraphics[width=0.48\linewidth]{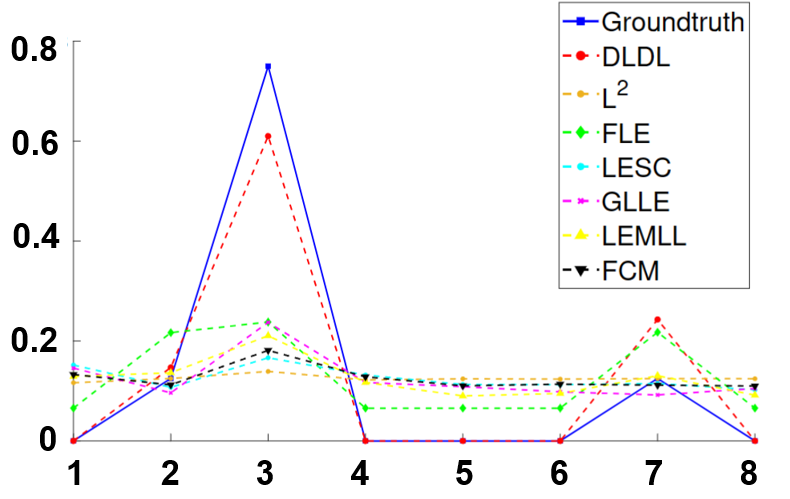}}
    \subfigure{
    \includegraphics[width=0.48\linewidth]{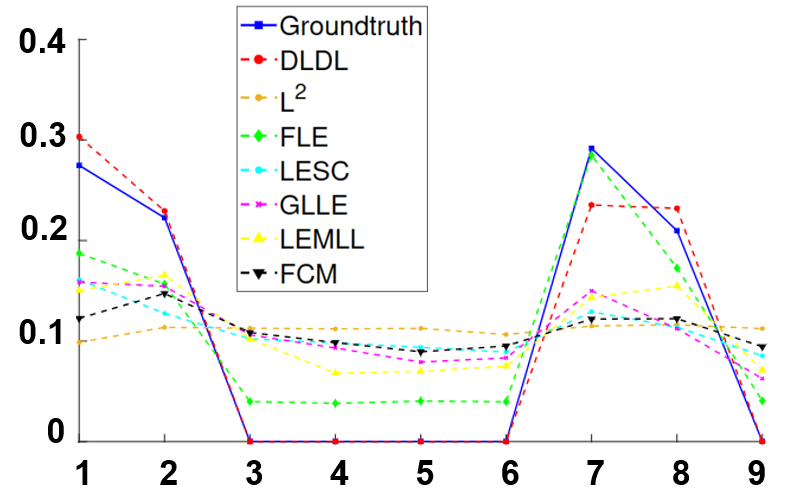}}
    \caption{The visualization of two typical recovery results on the Twitter (left) and NS (right) dataset.}
    \label{visualization}
\end{figure}

In this subsection, we present the visualization of two typical recovery results of all the models on two different datasets in Fig. \ref{visualization}. From it we can observe that DLDL has a better performance on the recovery of non-positive labels, and the recovery results of it are also more close to the ground-truth label distributions than the other comparison algorithms. For example, on the Twitter dataset, the logical values corresponding to the label 1, 4, 5, 6, 8 are all 0, indicating that these labels cannot describe this sample. However, the six comparison methods still assign positive description degrees for these invalid labels. Differently, DLDL avoids this problem, and produces differentiated description degrees fairly close to the ground-truth values.

\subsection{Results}

\subsubsection{Recovery Results}
In this subsection, we calculate the distances and similarities between the ground-truth LD and the LD recovered by each LE method via four evaluation metrics on each dataset. Table \ref{LEresults} shows the detailed recovery results, where all methods are ranked according to their performance and the best average rank on each metric is shown in boldface. In addition, the ``$\downarrow$" after the metrics means ``the smaller the greater", while the ``$\uparrow$" means ``the larger the greater". Based on the recovery results, we have the following observations:

\begin{itemize}
\item
DLDL achieves the lowest average rank in terms of all the four evaluation metrics. To be specific, out of the 24 statistical comparisons (6 datasets × 4 evaluation metrics), DLDL ranks 1st in all cases, which clearly shows the superiority of our approach in LE.

\item
On some datasets, our method achieves superior performance. For example, on FBP, REN and Twitter, DLDL significantly improves the results compared with other algorithms in terms of the Chebyshev, One-error and Intersection.

\iffalse
\item
Although the average ranks of FLE and GLLE are low, they do not perform well on all datasets as DLDL does. Specifically, FLE performs poorly on SCUT and RAF, and GLLE performs poorly on NS and REN. From this point of view, DLDL is more robust to different datasets than comparison algorithms.

\fi

\item
Under the metric of One-error, DLDL shows overwhelming advantages. 
This is attributed to the constraint $\textbf{0} \leq \boldsymbol{d}_i \leq \boldsymbol{y}_i$, which avoids assigning label description degrees to the negative logical labels. 

\end{itemize}

\begin{table*}[tb]
\centering
\resizebox{0.85\linewidth}{!}{
\begin{tblr}{
  cells = {c},
  cell{1}{1} = {r=2}{},
  cell{1}{2} = {c=6}{},
  cell{1}{8} = {r=2}{},
  cell{1}{9} = {c=6}{},
  cell{1}{15} = {r=2}{},
  cell{10}{1} = {r=2}{},
  cell{10}{2} = {c=6}{},
  cell{10}{8} = {r=2}{},
  cell{10}{9} = {c=6}{},
  cell{10}{15} = {r=2}{},
  vline{2,9} = {1-19}{},
  hline{1,3,5,10,12,14,19} = {-}{},
  hline{2,11} = {2-7,9-14}{},
}
Method  & Chebyshev$\downarrow$ &           &           &           &           &           & Avg.Rank & Clark$\downarrow$ &           &           &           &           &           & Avg.Rank \\
        & NS        & SCUT      & RAF       & FBP       & REN       & Twitter   &         & NS        & SCUT      & RAF       & FBP       & REN       & Twitter   &         \\
DLDL&\textbf{0.0845(1)}&\textbf{0.2821(1)}&\textbf{0.3133(1)}&\textbf{0.2783(1)}&\textbf{0.0306(1)}&\textbf{0.2989(1)}&\textbf{1.00(1)}&\textbf{2.3668(1)}&\textbf{0.9538(1)}&\textbf{1.0991(1)}&\textbf{1.0588(1)}&\textbf{0.8568(1)}&\textbf{1.2227(1)}&\textbf{1.00(1)}\\  
$\textit{L}^2$&0.3556(6)&0.3818(7)&0.3837(5)&0.3966(7)&0.6445(3)&0.5415(7)&5.83(6)&2.4620(5)&1.4968(7)&1.5966(2)&1.5060(6)&2.6508(3)&2.4016(7)&5.00(5)\\
FLE&0.3496(5)&0.3780(6)&0.3901(7)&0.3863(5)&0.6637(4)&0.3310(3)&5.00(5)&2.4682(6)&1.4949(6)&1.6153(7)&1.5011(5)&2.6574(4)&2.3846(6)&5.67(7)\\
GLLE&0.3257(2)&0.3466(2)&0.3801(3)&0.3630(2)&0.6686(5)&0.4710(4)&3.00(2)&2.4285(3)&1.4722(2)&1.6100(6)&1.4787(3)&2.6598(5)&2.3669(3)&3.67(3)\\
LEMLL&0.3291(3)&0.3527(3)&0.3699(2)&0.3897(6)&0.6369(2)&0.3271(2)&3.00(2)&2.4279(2)&1.4855(5)&1.6049(3)&1.6214(7)&2.6420(2)&2.3647(2)&3.50(2)\\
LESC&0.3601(7)&0.3665(5)&0.3845(6)&0.3790(4)&0.6746(7)&0.5105(6)&5.83(6)&2.4751(7)&1.4847(4)&1.6063(5)&1.4904(4)&2.6650(7)&2.3844(5)&5.33(6)\\
FCM&0.3466(4)&0.3596(4)&0.3825(4)&0.3638(3)&0.6725(6)&0.5064(5)&4.33(4)&2.4540(4)&1.4778(3)&1.6055(4)&1.4770(2)&2.6642(6)&2.3832(4)&3.83(4)\\

Method  & One-error$\downarrow$ &           &           &           &           &           & Avg.Rank & Intersection$\uparrow$ &           &           &           &           &           & Avg.Rank \\
        & NS        & SCUT      & RAF       & FBP       & REN       & Twitter   &         & NS        & SCUT      & RAF       & FBP       & REN       & Twitter   &         \\
DLDL&\textbf{0.0000(1)}&\textbf{0.0037(1)}&\textbf{0.0189(1)}&\textbf{0.0912(1)}&\textbf{0.0000(1)}&\textbf{0.0859(1)}&\textbf{1.00(1)}&\textbf{0.9082(1)}&\textbf{0.6894(1)}&\textbf{0.6298(1)}&\textbf{0.6987(1)}&\textbf{0.9694(1)}&\textbf{0.6985(1)}&\textbf{1.00(1)}\\  
$\textit{L}^2$&0.6119(3)&0.2743(6)&0.2810(2)&0.2765(4)&0.8157(7)&0.6578(6)&4.67(5)&0.4123(4)&0.5068(5)&0.4976(6)&0.5014(7)&0.2430(2)&0.3444(7)&5.17(6)\\
FLE&0.6364(6)&0.2693(5)&0.2904(7)&0.2754(3)&0.8130(6)&0.6569(3)&5.00(7)&0.3907(5)&0.5197(4)&0.4949(7)&0.5195(6)&0.2133(3)&0.5023(3)&4.67(4)\\
GLLE&0.6259(4)&0.2663(3)&0.2879(5)&0.2774(5)&0.8109(4)&0.6575(5)&4.33(3)&0.4586(3)&0.5623(2)&0.5077(3)&0.5466(2)&0.2046(4)&0.4264(4)&3.00(2)\\
LEMLL&0.6348(5)&0.2679(4)&0.2883(6)&0.3187(7)&0.7377(2)&0.6161(2)&4.33(3)&0.4739(2)&0.4635(6)&0.5280(2)&0.5368(4)&0.1826(7)&0.5866(2)&3.83(3)\\
LESC&0.6386(7)&0.2894(7)&0.2860(3)&0.2752(2)&0.8117(5)&0.6569(4)&4.67(5)&0.3885(6)&0.5350(3)&0.5005(5)&0.5228(5)&0.1948(7)&0.3726(6)&5.33(7)\\
FCM&0.5968(2)&0.2635(2)&0.2874(4)&0.2780(6)&0.7943(3)&0.6595(7)&4.00(2)&0.3724(7)&0.4319(7)&0.5011(4)&0.5437(3)&0.1852(2)&0.3826(5)&4.67(4)\\

\end{tblr}}

\caption{The recovery results of testing instances on the six datasets and the best average rank (i.e., Avg.Rank) is shown in boldface. The full table with standard deviation is provided in Section \textbf{D} in the supplementary file.}
\label{LEresults}
\end{table*}

\begin{table*}[tb]
\centering
\resizebox{0.85\linewidth}{!}{
\begin{tblr}{
  cells = {c},
  cell{1}{1} = {r=2}{},
  cell{1}{2} = {c=6}{},
  cell{1}{8} = {r=2}{},
  cell{1}{9} = {c=6}{},
  cell{1}{15} = {r=2}{},
  cell{11}{1} = {r=2}{},
  cell{11}{2} = {c=6}{},
  cell{11}{8} = {r=2}{},
  cell{11}{9} = {c=6}{},
  cell{11}{15} = {r=2}{},
  vline{2,9} = {1-20}{},
  hline{1,3,4,6,11,13,14,16,21} = {-}{},
  hline{2,12} = {2-7,9-14}{},
}
Method  & Chebyshev$\downarrow$ &           &           &           &           &           & Avg.Rank & Clark$\downarrow$ &           &           &           &           &           & Avg.Rank \\
        & NS        & SCUT      & RAF       & FBP       & REN       & Twitter   &         & NS        & SCUT      & RAF       & FBP       & REN       & Twitter   &         \\
SA-BFGS & 0.3533    & 0.4202    & 0.1583    & 0.3323    & 0.5871    & 0.3532    & -       &2.4212&1.5491&1.4531&1.4703&2.6407&2.5577&-\\
DLDL&0.4071(2)&0.4086(2)&\textbf{0.3911(1)}&\textbf{0.2972(1)}&\textbf{0.6164(1)}&\textbf{0.3657(1)}&\textbf{1.33(1)}&\textbf{2.5059(1)}&\textbf{1.5321(1)}&\textbf{1.6165(1)}&\textbf{1.1913(1)}&\textbf{2.6403(1)}&\textbf{2.5688(1)}&\textbf{1.00(1)}\\  
$\textit{L}^2$&0.4967(7)&0.4468(7)&0.4064(7)&0.4118(7)&0.6913(7)&0.5444(7)&7.00(7)&2.5111(4)&1.6470(7)&1.6356(7)&1.5491(7)&2.6723(6)&2.4084(3)&5.67(7)\\
FLE&\textbf{0.4011(1)}&0.4254(6)&0.3934(3)&0.3915(6)&0.6897(6)&0.4330(2)&4.00(5)&2.5062(2)&1.5426(3)&1.6246(5)&1.5013(5)&2.6735(7)&2.3951(4)&4.33(4)\\
GLLE&0.4257(3)&\textbf{0.4006(1)}&0.3968(4)&0.3641(3)&0.6833(4)&0.4807(4)&3.16(2)&2.5203(6)&1.6169(6)&1.6242(4)&1.4785(2)&2.6686(4)&2.3764(7)&4.83(6)\\
LEMLL&0.4490(4)&0.4148(5)&0.3975(5)&0.3286(2)&0.6321(2)&0.4706(3)&3.50(3)&2.5278(7)&1.5548(5)&1.6334(6)&1.5188(6)&2.6486(2)&2.4233(2)&4.67(5)\\
LESC&0.4743(6)&0.4143(4)&0.4008(6)&0.3795(5)&0.6873(5)&0.5138(6)&5.33(6)&2.5065(3)&1.5353(2)&1.6172(2)&1.4915(4)&2.6714(5)&2.3856(5)&3.50(2)\\
FCM&0.4692(5)&0.4132(3)&0.3931(2)&0.3689(4)&0.6736(3)&0.5071(5)&3.67(4)&2.5192(5)&1.5452(4)&1.6216(3)&1.4828(3)&2.6610(3)&2.3832(6)&4.00(3)\\

Method  & One-error$\downarrow$ &           &           &           &           &           & Avg.Rank & Intersection$\uparrow$ &           &           &           &           &           & Avg.Rank \\
        & NS        & SCUT      & RAF       & FBP       & REN       & Twitter   &         & NS        & SCUT      & RAF       & FBP       & REN       & Twitter   &         \\
SA-BFGS&0.4796&0.2940&0.2481&0.2718&0.7898&0.6075&-&0.5183&0.4731&0.8029&0.6226&0.3269&0.6035&-\\
DLDL&\textbf{0.5852(1)}&\textbf{0.2924(1)}&\textbf{0.2853(1)}&\textbf{0.1134(1)}&\textbf{0.1939(1)}&\textbf{0.3015(1)}&\textbf{1.00(1)}&0.4347(2)&\textbf{0.4825(1)}&\textbf{0.4864(1)}&\textbf{0.6523(1)}&\textbf{0.2719(1)}&\textbf{0.6025(1)}&\textbf{1.17(1)}\\  
$\textit{L}^2$&0.6549(7)&0.3404(7)&0.2909(4)&0.3171(7)&0.8149(4)&0.6605(7)&6.00(7)&0.3695(5)&0.3871(7)&0.4689(6)&0.5482(3)&0.1852(7)&0.3387(7)&5.83(7)\\
FLE&0.6525(6)&0.3093(4)&0.2931(6)&0.2754(2)&0.8159(5)&0.5791(3)&4.33(3)&0.3530(7)&0.4810(3)&0.4802(3)&0.5087(7)&0.1864(6)&0.3523(4)&5.00(6)\\
GLLE&0.6401(3)&0.3384(6)&0.2924(5)&0.2772(4)&0.8162(6)&0.6602(6)&5.00(6)&0.4002(3)&0.4250(6)&0.4643(5)&0.5455(4)&0.1974(4)&0.4117(2)&4.00(4)\\
LEMLL&0.6170(2)&0.3030(2)&0.2870(2)&0.2919(6)&0.8090(2)&0.5539(2)&2.67(2)&\textbf{0.4412(1)}&0.4514(5)&0.4776(4)&0.5929(2)&0.2636(2)&0.3392(6)&3.33(2)\\
LESC&0.6431(5)&0.3073(3)&0.2902(3)&0.2770(3)&0.8214(7)&0.6589(5)&4.33(3)&0.3772(4)&0.4679(4)&0.4703(5)&0.5238(6)&0.1910(5)&0.3770(3)&4.50(5)\\
FCM&0.6417(4)&0.3151(5)&0.2941(7)&0.2775(5)&0.8132(3)&0.6177(4)&4.67(5)&0.3583(6)&0.4818(2)&0.4824(2)&0.5389(5)&0.2070(3)&0.3408(5)&3.83(3)\\

\end{tblr}}

\caption{The predictive results of testing instances on the six datasets and the best average rank (i.e., Avg.Rank) is shown in boldface. The full table with standard deviation is provided in Section \textbf{D} in the supplementary file.}
\label{LDLresults}
\end{table*}

\subsubsection{Predictive Results}
In this subsection, SA-BFGS is used as the predictive model to generate the LDs of testing instances for the five LE methods, while DLDL and $\textit{L}^2$ can directly predict the LDs of testing instances. Then, we rank all methods and highlight the best average rank on each metric in boldface. The detailed predictive results are presented in Table \ref{LDLresults}. In addition, SA-BFGS is trained on the ground-truth LDs of the training instances and its results are recorded as the upper bound of the predictive results. From the reported results, we observe that:

\begin{itemize}
\item
DLDL achieves the lowest average rank in terms of the three evaluation metrics (i.e., Clark, Canberra and Intersection). Specifically, out of the 24 statistical comparisons, DLDL ranks 1st in 87.5\% cases and ranks 2nd in the remaining 12.5\% cases. In general, DLDL performs better than most comparison algorithms.

\item
In some cases, the performance of our approach even exceeds the upper bound. For example, on SCUT dataset, DLDL performs better than the SA-BFGS trained on the ground-truth LDs in terms of Chebyshev, Clark, Intersection and slightly on One-error, which shows that our model has considerable potential in learning from logical labels directly.

\item 
Although $\textit{L}^2$ is also a joint model, it doesn't consider the restriction of label distribution, and adopts an inappropriate fidelity term. Compared to $\textit{L}^2$, our method DLDL gets lower average ranks in terms of all the four metrics both in recovery and predictive results, indicating the effectiveness and superiority of our method.

\end{itemize}

\begin{table}[tb]
\Large

\centering
\renewcommand{\arraystretch}{1.6}
\resizebox{1\linewidth}{!}{
\begin{tabular}{c|cccccc|cccccc} 
\hline
\multirow{2}*{Method}&\multicolumn{6}{c|}{Recovery}&\multicolumn{6}{c}{Predictive}\\
\cline{2-13}
&{NS}&{SCUT}&{RAF}&{FBP}&{REN}&{Twitter}&{NS}&{SCUT}&{RAF}&{FBP}&{REN}&{Twitter}\\
\hline
DLDL&\textbf{0.1086}&\textbf{0.2854}&\textbf{0.3142}&\textbf{0.2771}&\textbf{0.0328}&\textbf{0.2091}&\textbf{0.3474}&\textbf{0.4021}&\textbf{0.3009}&\textbf{0.3113}&\textbf{0.6451}&\textbf{0.2622}\\
\hline
$\alpha=0$&0.2509&0.3072&0.3516&0.3687&0.0692&0.2092&0.4148&0.4207&0.3068&0.4867&0.7337&0.3515\\
$\beta=0$&0.1431&0.2943&0.3441&0.4254&0.0401&0.3026&0.4133&0.4104&0.3219&0.4730&0.6890&0.2911\\
$\gamma=0$&0.1087&0.3029&0.3357&0.3419&0.0400&0.3026&0.3649&0.4069&0.3115&0.4514&0.6733&0.3178\\
\hline
\end{tabular}}
\caption{Ablation study of recovery and predictive results (the selected metric is Chebyshev$\downarrow$)}
\label{LEablation}
\end{table}

\subsection{Further Analysis}
\subsubsection{Ablation Study}
In order to verify the necessity of the involved terms of our approach, we conduct ablation experiments and present the results on Chebyshev in Table \ref{LEablation}, where $\alpha$, $\beta$ and $\gamma$ are hyper-parameters to adjust the contributions of different terms. When one of hyper-parameters is fixed to 0, the remaining ones are determined by the performance on the validation set. From the results, we observe that the performance of DLDL becomes poor when the term controlled by $\beta$ is missing, indicating that it is critical to control the smoothness of the label distribution matrix $\mathbf{D}$. In general, DLDL outperforms its variants without some terms, which verify the effectiveness and rationality of our model.
\subsubsection{Significance Test}
In this subsection, a Bonferroni–Dunn test is applied to check whether DLDL 
achieves competitive performance compared to other algorithms. Through this test, we find that DLDL performs significantly better than the compared algorithms on the four metrics. For detailed significance test results, please refer to Section \textbf{E} in the supplementary file.

\section{Conclusion and Future Work}
This paper gives a preliminary positive answer to the question ``can we directly train an LDL model from the logical labels" \textit{both theoretically and empirically}. Specifically, we propose a new algorithm called DLDL, which unifies the conventional label enhancement and label distribution learning into a joint model. Moreover, our method avoids some common issues faced by the previous LE methods. Extensive experiments validate the advantage of DLDL against other LE algorithms in label enhancement, and also confirm the effectiveness of our method in directly training an LDL model from the logical labels. Nevertheless, our method is still inferior to the traditional LDL model when the ground-truth LD of the training set is available. In the future, we will explore possible ways to improve the predictive performance of our algorithm. 

\section*{Acknowledgements}

This work was supported by the National Natural Science Foundation of China under Grant (62106044) and the Natural Science Foundation of Jiangsu Province under Grant (BK20210221).

\bibliographystyle{named}
\bibliography{ijcai23}

\end{document}